# Software-Defined Robotics – Idea & Approach

Ali Al-Bayaty, *Member, IEEE*

*Abstract*— The methodology of Software-Defined Robotics hierarchical-based and stand-alone framework can be designed and implemented to program and control different sets of robots, regardless of their manufacturers' parameters and specifications, with unified commands and communications. This framework approach will increase the capability of (re)programming a specific group of robots during the runtime without affecting the others as desired in the critical missions and industrial operations, expand the shared bandwidth, enhance the reusability of code, leverage the computational processing power, decrease the unnecessary analyses of vast supplemental electrical components for each robot, as well as get advantages of the most state-of-the-art industrial trends in the cloud-based computing, Virtual Machines (VM), and Robot-as-a-Service (RaaS) technologies.

## I. INTRODUCTION

By decoupling the logical and the physical components apart, the economy of robotics will be shifted from the hardware manufacturing toward the software-based development trend. As a result, more innovations in software and potentially much more open robotic markets will result, and this will definitely reduce the non-recurring engineering costs [1]. "Software needs to decouple from hardware so developers have a common, wide-reaching marketplace to sell their solutions," said Nikhil Chauhan, Director of Marketing for GE Software, DevOpsAngle in 2013 [2].

By the year 2020, the 5G telecommunication infrastructures will enable robotics services, intelligent applications, different social networking websites, and digital economy to get benefits from such technology [3]. Cloud computing, Software-Defined Networking (SDN) [4] [5], Virtual Machines (VM) [6] [7], and vast of networking virtualizations are new concepts in networking, but they are all just different views of telecommunications infrastructure *softwarization*. Such developing trend will definitely speed up the telecommunications innovation by increasing its programmability through the APIs (Application Programming Interfaces), optimizing costs and reducing time-to-market factor, and also providing better new benefits and services as well [8].

So, what should be happened when mixing cloud-based computing and storage technologies with the robotics field? Robots become more powerful and smarter, capable of communicating and collaborating with each other, as well as learning from each other's mistakes; thus, they can accomplish a variety of tasks in more effective ways. "The problem is that in many cases, the different physical hardware of each robot limits the applicability of other robot's shared knowledge," said Christian Penaloza, a researcher at Advanced Telecommunications Research (ATR) Institute [9].

So, designing, developing, and managing large numbers of different types of robots, e.g. MAS (Multi-Agent System), swarm robots, or search-and-rescue robots, can raise the dilemma of resource consuming during the operational missions, such as time wasting for cloning similar codes many times, learning new and specific programming languages, limiting the shared bandwidth, using fixed processing power, and an unnecessary analyzing of different supplemental electrical components for each robot. With the diversity of manufacturers' specifications and private parameters, the difficulty of programming and managing robots individually appears when (re-)configuring these robots to achieve a required task or a desired formation. This dilemma can frequently occur in the industrial and the operational fields; for that, a dynamic configuration system for a group of robots is needed and depended upon their critical updates in the mission without the need to restart it. Thus, a unified framework model is in need to solve these difficulties with a more systematic approach.

## II. SOFTWARE-DEFINED ROBOTICS FRAMEWORK

Robotics hierarchical-based framework can be designed upon the similarity of the working principles of the SDN that allows network administrators to control and manage the networking entities through higher-level functional abstraction with the aid of OpenFlow protocol; hence, robots and other sensory devices will be a replacement for these networking entities. The proposed name for this framework is *Software-Defined Robotics* (SDBotics), and the proposed communication protocol that has as similar philosophy as OpenFlow protocol is termed *OpenBots*. Thus, OpenBots is a communication protocol that sends the configurations and the operational controls as unified mnemonics to a distinct robot, a group of robots, or all of them through the SDBotics framework.

The SDBotics framework consists of three layers, which are (from top to bottom, and termed as *ACE*): *Applications*, *Controller*, and *Entities*. The Applications layer is the interfacing layer for the user, who develops and manages the robot(s), using different sets of GUI-based (Graphical User Interface) and CLI-based (Command Line Interface) programming languages or VPLs (Visual Programming Languages) that are compatible with the SDBotics framework through pre-defined APIs and libraries.

Ali Al-Bayaty is with the Department of Electrical and Computer Engineering, University of New Haven, West Haven, CT 06516 USA (corresponding author to provide phone: 203-936-9559; e-mail: aalba3@unh.newhaven.edu).

The Controller layer is responsible for:

- Managing the robots' topology or formation.
- Providing statistical information regarding the overall networking performance.
- Tracking each robot's parameters, and providing global localization and mapping.
- Finding the shortest communication path.
- Traffic redirecting and forwarding of the network packets, and proving other networking services, such as affinity, LISP (Locator/Identifier Separation Protocol), DDoS (Distributed Denial of Service) protection, etc.
- Receiving the unified mnemonics from the above layer (Applications layer) using REST (Representational State Transfer) web service protocol through the Northbound interfaces, for further analyzing, debugging, and judging issues.
- Sending the unified mnemonics to the below layer (Entities layer) using OpenBots protocol through the Southbound interfaces to be broadcasted to the designated robot(s).

Whereas, the Entities layer consists of robots and other sensory devices that receive the unified mnemonics and configurations from the user(s) in the Applications layer. And, Fig. 1 demonstrates the hierarchical layers of the SDBotics framework.

Finally, each robot has its own separate buffer (like the flow-table entries of the OpenFlow-enabled switches in the SDN) that stores the received mnemonics from the Controller layer through the OpenBots protocol, and then translates these unified mnemonics to their equivalent manufacturer's instructions. Figure 2 illustrates the structure of the OpenBots protocol.

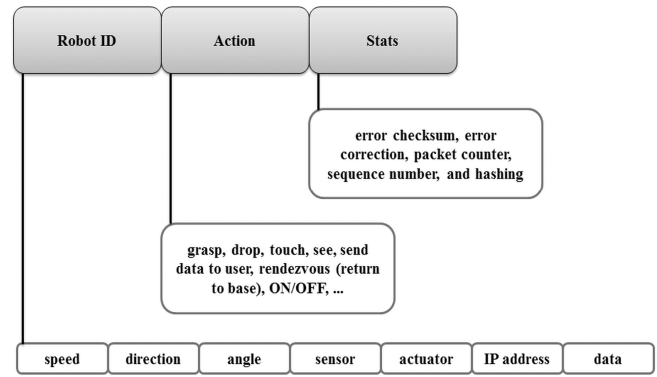

Figure 2. The OpenBots protocol structure.

Moreover, the Controller layer can be implemented in two approaches: as a centralized approach by embedding it in one of the robots or as a cloud-based approach by deploying it on a server over the Internet. The first approach, as illustrated in Fig. 3 (a), can be used in a mission that requires a small number of robots, limited bandwidth, and lower processing capability to achieve the overall task. In contrast, the second approach, as shown in Fig. 3 (b), can be implemented for huge and vast tasks that a single robot holds the Controller layer can't capable of achieving them, due to a large number of used sensory data and high computational processing demands.

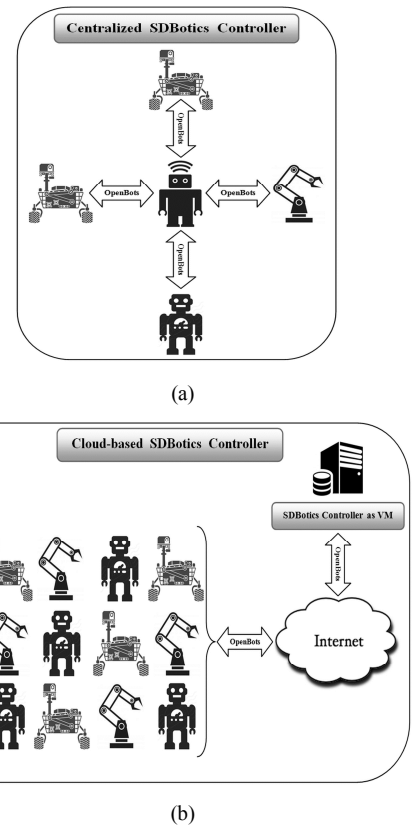

Figure 3. The SDBotics Controller layer:
(a) Centralized approach, and
(b) Cloud-based approach.

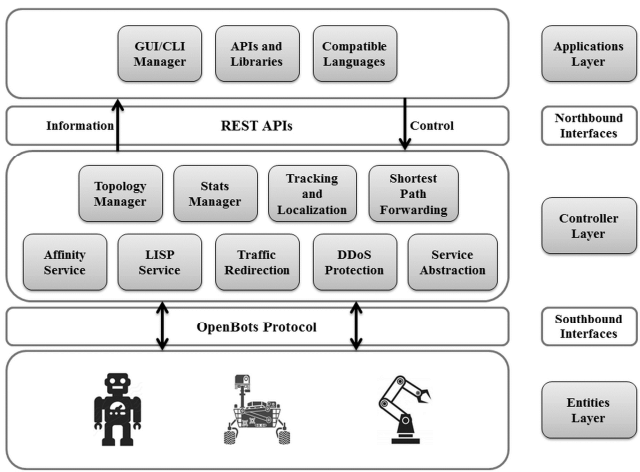

Figure 1. The hierarchical layers of the SDBotics framework.

## A. Framework Representation Using Sets

Assuming a robot (`R_i`) in the Entities layer has a set of coefficients that are the programmable configurations and parameters of the `R_i` itself, like speed, direction, angle, sensor, actuator, IP address, and data, as represented in (1).

```
R_i = {i, speed, dir, angle, sensor,
       actuator, ip-addr, data}         (1)
```

Where:

- `i`: The robot's identification number.
- `speed`: 1 as stop, 2 as normal speed, 3 as accelerated speed, …
- `dir`: 1 as forward and 2 as backward movement directions.
- `angle`: Rotational angles from $\angle 0^o$ to $\angle 180^o$.
- `sensor`: 1 as touch sensor, 2 as proximity sensor, 3 as camera, …
- `actuator`: 1 as gripper, 2 as camera-mounted motor, …
- `ip-addr`: IPv4 or IPv6 address.
- `data`: Bi-directional data from/to robot/user.

The packet structure of the OpenBots protocol (`O_j`) consists of a set of elements that are transmitted among the robots and the Controller layer, as stated in (2).

```
O_j = {R_i, action, stats}              (2)
```

Where,

- `action`: The specific task, e.g. grasp, drop, touch, see, send data to the user, Rendezvous (return to the base), ON/OFF, ...
- `stats`: The statistical status of `O_j` packet itself, e.g. error checksum, error correction, packet counter, sequence number, and other security features, like hashing.

Hence, ($R_i \in O_j$).

Therefore, the Controller layer (`S`) will have a large set of `O_j`'s, depending on the available number of robots in the overall framework, as represented in (3).

```
S = {O_1, O_2, … , O_j}                 (3)
```

For that,

∵ $R_i \subseteq O_j$ and $O_j \in S$

∴ $R_i \subseteq S$

As a result, each newly joined robot (`R_i`) in the SDBotics framework should be implied to `S`; and each ($R_i \cup O_j$) will form a completed communication linkage between a robot (Entities layer) and the Controller layer in the overall `S`, as stated in (4).

$$\therefore \forall\ (R_i \cup O_j) \subseteq S \qquad (4)$$

Where,

i = j = 1, 2, 3, …

And, Fig. 4 illustrates the sets representation of the SDBotics framework.

## B. Interpretive Example

In order to interpret the complete scenario of the overall SDBotics workflow, suppose a user wants to perform the following tasks for the robot (`R_3`) using the Applications layer's compatible programming language, as follows:

- Turn 'on' $R_3$,
- Move 'forward' with a 'normal' speed,
- When an object is 'touched', 'stop', and 'grasp' it,
- Return $R_3$ back, with the picked object, to the 'same starting point',
- 'Drop' the object,
- 'Send' the statement 'DONE' to the user, and finally
- Turn 'off' $R_3$.

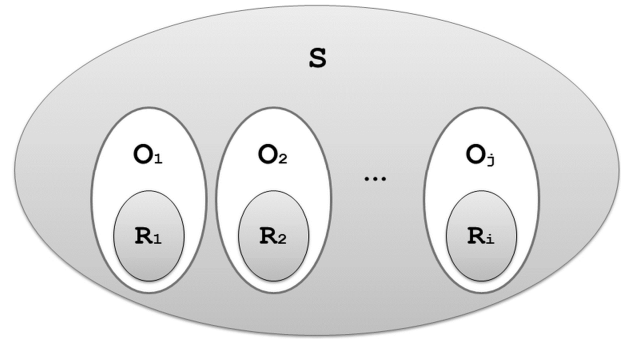

Figure 4. The Sets representation of the SDBotics framework.

For simplicity, the previously mentioned tasks are coded using the pseudo-code format, as the following code:

```
// -----------------------------------------
// SDBotics Applications Layer's Compatible
// Programming Language:
// -----------------------------------------

// Importing SDBotics APIs and libraries:
IMPORT <SDBotics-library>

// -----------------------------------------
// Declaring robots' coefficients & actions,
// the SDBotics framework will populate these
// coefficients and actions as unified
// mnemonics to the designated robot(s) in
// the Entities layer:
// -----------------------------------------

SDBotics->Init()

LOOP
(
  SDBotics->PopulateData() =
  [
    // ('robotID', speed, dir, angle, sensor,
    // actuator, 'ip-addr', 'data', 'action')

    ('R3',2,1,0,1,1,'192.168.0.3','','ON');
    ('R3',1,1,0,1,1,'192.168.0.3','','TOUCH');
    ('R3',1,1,0,1,1,'192.168.0.3','','GRASP');
    ('R3',2,1,180,1,1,'192.168.0.3','',
                                  'RENDEZVOUS');
    ('R3',1,1,0,1,1,'192.168.0.3','','DROP');
    ('R3',1,1,0,1,1,'192.168.0.3','DONE',
                                  'SEND');
    ('R3',1,1,0,1,1,'192.168.0.3','','OFF');

    ('xxx', …);   // Other robots' coefficients
                  // & actions.
  ]
)
```

Note that, the proposed programming language for the Applications layer is as similar as the functional language model that supports commands abstraction, so a user or a developer will not be aware of the technical and the hardware details for each robot in the SDBotics framework. The Controller layer will take these mnemonics from the Applications layer and populate them to the specific robots in the Entities layer through the OpenBots protocol, and then these robots will store these mnemonics in their buffers for decoding and translating to their equivalent manufacturers' commands and parameters. Further development to the Applications Layer's compatible programming software can be done and improved by using robotics visual representational languages, e.g. VPLs.

## III. CONCLUSION

By designing and implementing the SDBotics hierarchical-based and stand-alone framework, this will inspire different approaches in robotics regarding the restructuring of formations and self-governing, create independent configurations on the runtime and apart from the restricted manufacturers and vendor-specific parameters and conditions, be totally isolated from the traditional SDN frameworks and their requirements, produce more open-hardware markets and industries, and integrate the current cloud-based trends, like VM and Robot-as-a-Service (RaaS) [10] [11], in more innovative ways.

With further development, compatible APIs, pre-defined libraries, and programming software can be developed and enhanced to support the Applications layer and pace the robotics programming in the SDBotics framework seamlessly. For distributed applications, fail-safe techniques, and with the aid of the 5G technology, the SDBotics framework with all of its hierarchal layers and supported protocols can be deployed on smartphones, due to their recent efficient hardware, stable operating systems, and the availability of different sensors and communication modules.

Finally, non-professional users will be able to develop and program different types of robots without knowing the technical details of the used hardware for each robot; together, the SDBotics framework and the OpenBots protocol can be used as academic tools and as open-source projects for universities and scientific communities, which will, in turn, simplify the different strategies of teaching students in robotics.